\documentclass[sectrefs]{svmult}

\usepackage{mathptmx}       
\usepackage{helvet}         
\usepackage{courier}        
\usepackage{type1cm}        

\usepackage{makeidx}         
\usepackage{graphicx}        
\usepackage{multicol}        
\usepackage[bottom]{footmisc}

\usepackage{newtxtext}       %
\usepackage{newtxmath}       
\usepackage{booktabs}
\usepackage{url}

\newcommand{\etal}{\textit{et al.}}

\makeindex             

\begin{document}
\title*{Advances in Deep Learning for Hyperspectral Image Analysis -- Addressing Challenges Arising in Practical Imaging Scenarios}
\label{advanDL} 
\author{Xiong Zhou and Saurabh Prasad}
\institute{Xiong Zhou \at Amazon Web Service, AI, Seattle USA \email{xiongzho@amazon.com}
	\and Saurabh Prasad \at Hyperspectral Image Analysis Group, Department
	of Electrical and Computer Engineering, University of Houston, Houston USA \email{saurabh.prasad@ieee.org}}
\maketitle

\abstract{Deep neural networks have proven to be very effective for computer vision tasks, such as image classification, object detection, and semantic segmentation -- these are primarily applied to color imagery and video. In recent years, there has been an emergence of deep learning algorithms being applied to hyperspectral and multispectral imagery for remote sensing and biomedicine tasks. These multi-channel images come with their own unique set of challenges that must be addressed for effective image analysis. Challenges include limited ground truth (annotation is expensive and extensive labeling is often not feasible), and high dimensional nature of the data (each pixel is represented by hundreds of spectral bands), despite being presented by a large amount of unlabeled data and the potential to leverage multiple sensors/sources that observe the same scene. In this chapter, we will review recent advances in the community that leverage deep learning for robust hyperspectral image analysis despite these unique challenges -- specifically, we will review unsupervised, semi-supervised and active learning approaches to image analysis, as well as transfer learning approaches for multi-source (e.g. multi-sensor, or multi-temporal) image analysis.}

\section{Deep Learning -- Challenges presented by Hyperspectral Imagery}
\label{advanDL-why}
Since AlexNet~\cite{krizhevsky2012imagenet} won the ImageNet challenge in 2012, deep learning approaches have gradually replaced traditional methods becoming a predominant tool in a variety of computer vision applications. Researchers have reported remarkable results with deep neural networks in visual analysis tasks such as image classification, object detection, and semantic segmentation. A major differentiating factor that separates deep learning from conventional neural network based learning is the amount of parameters in a model. With hundreds of thousands even millions or billions of parameters, deep neural networks use techniques such as error backpropagation~\cite{rumelhart1988learning}, weight decay~\cite{krogh1992simple}, pretraining~\cite{hinton2006fast}, dropout~\cite{srivastava2014dropout}, and batch normalization~\cite{ioffe2015batch} to prevent the model from overfitting or simply memorizing the data. Combined with the increased computing power and specially designed hardware such as Graphics Processing Units (GPU), deep neural networks are able to learn from and process  unprecedented large-scale data to generate abstract yet discriminative features and classify them.


Although there is a significant potential to leverage from deep learning advances for hyperspectral image analysis, such data come with unique challenges which must be addressed in the context of deep neural networks for effective analysis. It is well understood that deep neural networks are notoriously data hungry insofar as training the models is concerned. This is attributed to the manner in which neural networks are trained. A typical training of a network comprises of two steps: 1) pass data through the network and compute a task dependent loss; and 2) minimize the loss by adjusting the network weights by back-propagating the error~\cite{rumelhart1988learning}. During such a process, a model could easily end up overfitting~\cite{caruana2001overfitting}, particularly if we do not provide sufficient training data. Data annotation has always been a major obstacle in machine learning research -- and this requirement is amplified with deep neural networks. Acquiring extensive libraries such as ImageNet~\cite{deng2009imagenet} for various applications may be very costly and time consuming. 
This problem becomes even more acute when working with hyperspectral imagery for applications to remote sensing and biomedicine. Not only does one need specific domain expertise to label the imagery, annotation itself is challenging due to the resolution, scale and interpretability of the imagery even by domain experts. 
For example, it can be argued that it is much more difficult to tell the different types of soil tillage apart by looking at a hyperspectral image than discerning everyday objects in color imagery. Further, the ``gold-standard'' in annotating remotely sensed imagery would be through field campaigns where domain experts verify the objects at exact geolocations corresponding to the pixels in the image. This can be very time consuming and for many appplications unfeasible. It is hence common in hyperspectral image analysis tasks to have a very small set of labeled ground truth data to train models from. 

In addition to the label scarcity, the large inter-class variance of hyperspectral data also increases the complexity of the underlying classification task. Given the same material or object, the spectral reflectance (or absorbance) profiles from two hyperspectral sensors could be dramatically different because of the differences in wavelength range and spectral resolution. Even when the same sensor is used to collect images, one can get significant spectral variability due to the variation of view angle, atmospheric conditions, sensor altitude, geometric distortions etc.~\cite{camps2014advances}. {Another reason for high spectral variability is mixed pixels arising from imaging platforms that result in low spatial resolution -- as a result, the spectra of one pixel corresponds to more than one object on the ground \cite{bioucas2012hyperspectral}. }

For robust machine learning and image analysis, there are two essential components -- deploying an appropriate machine learning model, and leveraging a library of training data that is representative of the underlying inter-class and intra-class variability. For Image analysis, specifically for classification tasks, deep learning models are variations of convolution neural networks (CNNs)~\cite{lecun2010convolutional}, which conduct a series of 2D convolutions between input images and (spatial) filters in a hierarchical fashion. It has been shown that such hierarchical representations are very efficient in recognizing objects in natural images~\cite{boureau2010learning}. 
When working with hyperspectral images, however, CNN based features ~\cite{yosinski2015understanding} such as color blobs, edges, shapes etc. may not be the only features of interest for the underlying analysis. There is important information encoded in the spectral profile which can be very helpful for analysis. 
Unfortunately, in traditional applications of CNNs to hyperspectral imagery, modeling of spectral content in conjunction with spatial content is ignored. Although one can argue that spectral information could still be picked up when 2D convolutions are applied channel by channel or features from different channels are stacked together, such approaches would not constitute optimal modeling of spectral reflectance/absorbance characteristics. It is well understood that when the spectral correlations are explicitly exploited, spectral-spatial features are more discriminative -- from traditional wavelet based feature extraction ~\cite{shen2011three,zhou2016wavelet} to modern CNNs~\cite{chen2016deep,li2017spectral,paoletti2018new,zhong2018spectral}. In chapters 3 and 4, we have reviewed variations of convolutional and recurrent neural networks that model the spatial and spectral properties of hyperspectral data. In this chapter, we review recent works that specific address issues arising from deploying deep learning neural networks in challenging scenarios. In particular, our emphasis is on challenges presented by (1) limited labeled data, wherein one must leverage the vast amount of available unlabeled data in conjunction with limited data for robust learning, and (2) multi-source optical data, wherein it is important to transfer models learned from one source (e.g. a specific sensor/platform/viewpoint/timepoint), and transfer the learned model to a different source (a different sensor/platform/viewpoint/timepoint), with the assumption that one source is rich in the quality and/or quantity of labeled training data while the other source is not.  

\section{Robust Learning with Limited Labeled Data}
\label{advanDL-train}
To address the labeled data scarcity, one strategy is to recruit resources (time and money, for example) with the goal of expanding the training library by annotating more data. However, for many applications, human annotation is neither scalable nor sustainable. 
An alternate (and more practical) strategy to address this problem is to design algorithms that do not require a large library of training data, but can instead learn from the extensive unlabeled data in conjunction with the limited amount of labeled data. Within this broad theme, we will review unsupervised feature learning, semi-supervised and active learning strategies. {We will present results of several methods discussed in this chapter with three hyperspectral datasets - two of these are benchmark hyperspectral datasets, University of Pavia~\cite{paviau} and University of Houston~\cite{uhdata}, and represent urban land-cover classification tasks. The University of Pavia dataset is a hyperspectral scene representing 9 urban land cover classes, with 103 spectral channels spanning the visible through near-infrared region. The 2013 University of Houston dataset is a hyperspectral scene acquired over the University of Houston campus, and representing 15 urban land cover classes. It has 144 spectral channels in the visible through near infrared region. 
	The third dataset is a challenging multi-source (multi-sensor/multi-viewpoint) hyperspectral dataset~\cite{zhou2017domain} that is particularly relevant in a transfer learning context -- details of this dataset are presented later in section \ref{advanDL-trans}.}

\subsection{Unsupervised Feature Learning}
\label{advanDL-us}
In contrast to the labeled data, unlabeled data are often easy and cheap to acquire for many applications, including remotely sensed hyperspectral imagery. Unsupervised learning techniques do not rely on labels, and that makes this class of methods very appealing. Compared to supervised learning where labeled data are used as a ``teacher'' for guidance, models trained with unsupervised learning tend to learn relationships between data samples and estimate the data properties class-specific labelings of samples. In the sense that most deep networks can be comprised of two components - a feature extraction front-end, and an analysis backend (e.g. undertaking tasks such as classification, regression etc.), an approach can be completely unsupervised relative to the training labels (e.g. a neural network tasked with fusing sensors for super-resolution), or completely supervised (e.g. a neural network wherein both the features and the backend classifiers are learned with the end goal of maximizing inter-class discrimination. There are also scenarios wherein the feature extraction part of the network is unsupervised (where the labeled data are not used to train model parameters), but the backend (e.g. classification) component of the network is supervised. In this chapter, whenever the feature extraction component of a network is unsupervised (whether the backend model is supervised or unsupervised), we refer to this class of methods as carrying out ``unsupervised feature learning''.

The benefit of unsupervised feature learning is that we can learn useful features (in an unsupervised fashion) from a large amount of unlabeled data (e.g. spatial features representing the natural characteristics of a scene) despite not having sufficient labeled data to learn object-specific features, with the assumption that the features learned in an unsupervised manner can still positively impact a downstream supervised learning task.

In traditional feature learning (e.g. dimensionality reduction, subspace learning or spatial feature extraction), the processing operators are often based on assumptions or prior knowledge about data characteristics. Optimization of feature learning to a task at hand is hence non-trivial. Deep learning-based methods address this problem in a data-adaptive manner, where the feature learning is undertaken in the context of the overall analysis task in the same network. 

Deep learning-based strategies, such as autoencoders ~\cite{rumelhart1985learning} and their variants, restricted Boltzmann machines (RBM)~\cite{smolensky1986information,hinton2002training}, and deep belief networks (DBN)~\cite{hinton2006reducing}, have exhibited a potential to effectively characterize hyperspectral data. For classification tasks, the most common way to use unsupervised feature learning is to extract (\textit{learn}) features from the raw data that can then be used to train classifiers downstream. 
Section 3.1 in Chapter 3 describes such a use of autoencoders for extracting features for tasks such as classification.

In Chen \etal\ ~\cite{chen2014deep}, the effectiveness of autoencoder derived features was demonstrated for hyperspectral image analysis. Although they attempted to incorporate spatial information by feeding the autoencoder with image patches, a significant amount of information is potentially lost due to the flattening process. To capture the multi-scale nature of objects in remotely sensed images, image patches with different sizes were used as inputs for a stacked sparse autoencoder in~\cite{tao2015unsupervised}. To extract similar multi-scale spatial-spectral information, Zhao \etal\ ~\cite{zhao2017spectral} applied a scale transformation by upsampling the input images before sending them to the stacked sparse autoencoder. Instead of manipulating the spatial size of inputs, Ma \etal\ ~\cite{ma2016spectral} proposed to enforce the local constraint as a regularization term in the energy function of the autoencoder. By using a stacked denoising autoencoder, Xing \etal\ ~\cite{xing2016stacked} sought to improve the feature stability and robustness with partially corrupted inputs. Although these approaches have been effective, they still require input signals (frames/patches) to be reshaped as one dimensional vectors, which inevitably results in a loss of spatial information. To better leverage the spatial correlations between adjacent pixels, several works have proposed to use the convolutional autoencoder to extract features from hyperspectral data~\cite{kemker2017self,ji2017learning,han2017spatial}.

Stacking layers has been shown to be an effective way to increase the representation power of an autoencoder model. The same principle applies to  deep belief networks ~\cite{le2008representational}, where each layer is represented by a restricted Boltzmann machine. With the ability to extract a hierarchical representation from the training data, promising results have been shown for DBN/RBM for hyperspectral image analysis~\cite{li2014classification,midhun2014deep,chen2015spectral,tao2017unsupervised,zhou2017deep,li2019deep,tan2019parallel}. In recent works, some alternate strategies to unsupervised feature learning for hyperspectral image analysis have also emerged. In~\cite{romero2016unsupervised}, a convolutional neural networks was trained in a greedy layer-wise unsupervised fashion. A special learning criteria called enforcing population and lifetime sparsity (EPLS)~\cite{romero2015meta} was utilized to ensure that the generated features are unique, sparse and robust at the same time. In~\cite{haut2018new}, the hourglass network~\cite{newell2016stacked}, which shares a similar network architecture as an autoencoder, was trained for super-resolution using unlabeled samples in conjunction with noise. The reconstructed image was downsampled and compared with the real low-resolution image. The offset between these two was used as the loss function that was minimized to train the entire network. A minimized loss (offset) indicates the reconstruction from the network would be a good super-resolved estimate of the original image.

\subsection{Semi-supervised learning}
\label{advanDL-ss}
Although the feature learning strategy allows us to extract informative features from unlabeled data, the classification part of the network still requires labeled training samples. Methods that rely completely on unsupervised learning may not provide discriminative features from unlabeled data entirely for challenging classification tasks. Semi-supervised deep learning is an alternate approach where unlabeled data are used in conjunction with a small amount of labeled data to train deep networks (both the feature extraction and classification components). It falls between supervised learning and unsupervised learning and leverages benefits of both approaches. In the context of classification, semi-supervised learning often provides better performance compared to unsupervised feature learning, but without the annotation/labeling cost needed for fully supervised learning~\cite{chapelle2009semi}. 

Semi-supervised learning has been shown to be beneficial for hyperspectral image classification in various scenarios~\cite{liu2017semi,he2017generative,kemker2018low,niu2018weakly,wu2018semi,kang2019semi}. Recent research~\cite{kemker2018low} has shown that the classification performance of a multilayer perceptron (MLP) can be improved by adding an unsupervised loss. 
In addition to the categorical cross entropy loss, a symmetric decoder branch was added to the MLP and multiple reconstruction losses, {measured by the mean squared error of the encoder and decoder}, were enforced to help the network generate effective features.
The reconstruction loss in fact served as a regularizer to prevent the model from overfitting. A similar strategy has been used with convolutional neural networks in~\cite{liu2017semi}.

\begin{figure}[h]
	\centering
	\includegraphics[width=0.6\textwidth]{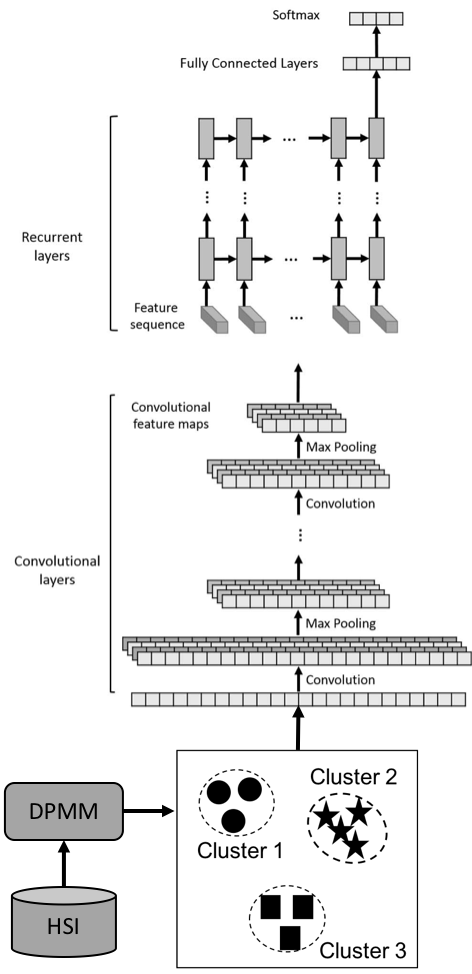}
	\caption{Architecture of the convolutional recurrent neural network. Cluster labels are used for pretraining the network. (Source adapted from~\cite{wu2018semi})}
	\label{fig:advanDL-crnn}
\end{figure}

A variant of semi-supervised deep learning, proposed by Wu and Prasad in~\cite{wu2018semi} entails learning a deep network that extracts features that are discriminative from the perspective of the intrinsic clustering structure of data (i.e., these deep features can discriminate between cluster labels -- also referred to as pseudo-labels in this work) -- in short, the cluster labels generated from clustering of unlabeled data can be used to boost the classification performance. To this end, a constrained Dirichlet Process Mixture Model {(DPMM)} was used, and a variational inference scheme was proposed to learn the underlying clustering from data. The clustering labels of the data were used as \emph{pseudo labels} for training a convolutional recurrent neural network, where a CNN was followed by a few recurrent layers (akin to a pretraining with pseudo-labels). Figure~\ref{fig:advanDL-crnn} depicts the architecture of network. {The network configuration is specified in Table~\ref{tab:advanDL-crnn}, where convolutional layers are denoted as ``conv <filter size> - <number of filters> and recurrent layers are denoted as ``recur-<feature dimension>''.}
\begin{table}[h] 
	\centering
	\caption{Network configuration summary for the Aerial view wetland hyperspectral dataset. Every convolutional layer is followed by a max pooling layer, which is omitted for the sake of simplicity. (Source adapted from \cite{wu2018semi})}
	\begin{tabular}{c}
		\hline
		input-103 $\rightarrow$ conv3-32 $\rightarrow$ conv3-32 $\rightarrow$ conv3-64 $\rightarrow$ conv3-64 \\
		$\rightarrow$ recur-256 $\rightarrow$ recur-512 $\rightarrow$ fc-64 $\rightarrow$ fc-64 $\rightarrow$ softmax-9 	\\
		\hline
	\end{tabular}
	\label{tab:advanDL-crnn}
\end{table}

After pretraining with unlabeled data and associated pseudo-labels, the network was fine-tuned with labeled data. This entails adding a few more layers to the previously trained network and learning only these layers from the labeled data. {Compared to traditional semi-supervised methods, the pseudo-label-based network, PL-SSDL, achieved higher accuracy on the wetland data {(a detailed description of this dataset is provided in Section~\ref{advanDL-trans})} as shown in Table~\ref{tab:advanDL-crnn-acc}. The effect of varying depth of the pretrained network on the classification performance is shown as Fig.~\ref{fig:advanDL-crnn-depth}. Accuracy increases as the model goes deeper, i.e., more layers.} {In addition to the environmental monitoring application represented by the wetland dataset, the efficacy of PL-SSDL was also verified for urban land-cover classification tasks using the University of Pavia~\cite{paviau} and the University of Houston~{\cite{uhdata} datasets, having 9 and 15 land cover classes, and representing spectra 103 and 144 spectral channels spanning the visible through near infrared regions respectively.} As we can see from Figure~\ref{fig:advanDL-plssdl}, features extracted with pseudo label (middle column) are separated better than the raw hyperspectral data (left column), which implies pretraining with unlabeled data makes the features more discriminative. Compared to a network that is trained solely using labeled data, the semi-supervised method requires much less labeled samples due to the pretrained model. With only a few labeled samples per class, features are further improved by fine-tuning (right column) the network. Similar to this idea, Kang~\cite{kang2019semi} later trained a CNN with pseudo labels to extract spatial deep features through pretraining.
	\begin{table}[h]
		\centering
		\begin{tabular}{cccccc}
			\toprule
			Methods & Label propagation & TSVM & SS-LapSVM & Ladder Networks & PL-SSDL \\
			\midrule
			Accuracy & $89.28 \pm 1.04$ & $92.24 \pm 0.81$ & $95.17 \pm 0.85$ & $93.17 \pm 1.49$ & $97.33 \pm 0.48$ \\
			\bottomrule
		\end{tabular}
		\caption{Overall classification accuracies of different methods on the aerial view wetland dataset.  (Source adapted from \cite{wu2018semi})}
		\label{tab:advanDL-crnn-acc}
	\end{table}
	
	\begin{figure}[h]
		\centering
		\includegraphics[width=0.75\textwidth]{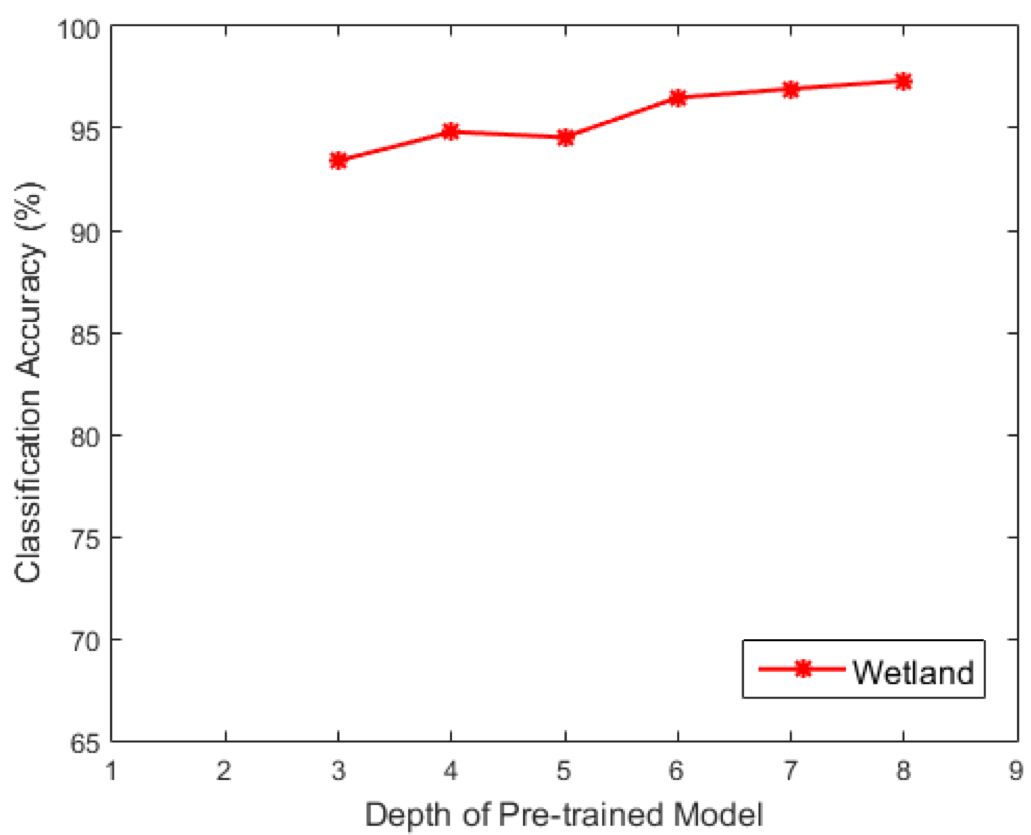}
		\caption{Classification accuracy as a function of the depth of the pretrained model. (Source adapted from \cite{wu2018semi})}
		\label{fig:advanDL-crnn-depth}
	\end{figure}
	\begin{figure}[h]
		\centering
		\includegraphics[width=\textwidth]{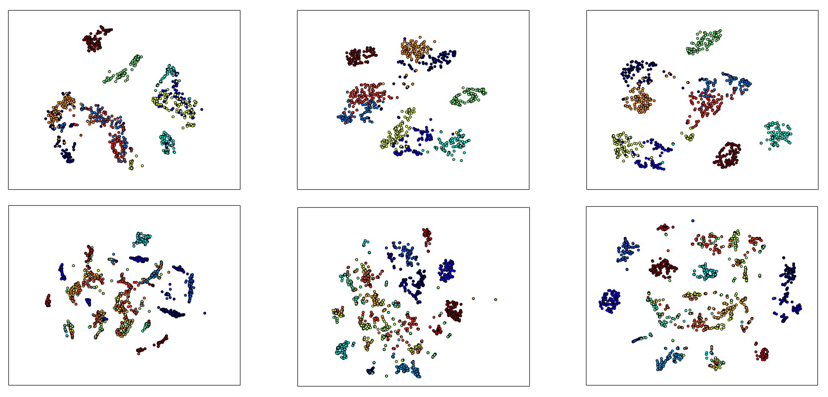}
		\caption{t-SNE visualization of features at different training stages on the University of Pavia~\cite{paviau} (top row) and University of Houston~\cite{uhdata} (bottom row) datasets. Left column represents raw image features, middle column represents features after unlabeled data pretraining, and right column represents feature after labeled data fine-tuning.  (Source adapted from \cite{wu2018semi})}
		\label{fig:advanDL-plssdl}
	\end{figure}
	
	\subsection{Active learning}
	\label{advanDL-al}
	Leveraging unlabeled data is the underlying principle of unsupervised and semi-supervised learning. Active learning, on the other hand, aims to improve the efficiency of acquiring labeling data as much as possible. Figure~\ref{fig:advanDL-al} shows a typical active learning flow, which contains four components: a labeled training set, a machine learning model, an unlabeled pool of data, and an oracle (a human annotator / domain expert). The labeled set is initially used for training the model. Based on the model's prediction, queries are then selected from the unlabeled pool and sent to the oracle for labeling. The loop is iterated until a pre-determined convergence criterion is met. The criteria used for selecting samples to query determines the efficiency of model training -- efficiency here refers to the machine learning model reaching its full discriminative potential using as few queried labeled samples as possible. If every queried sample provides significant information to the model when labeled and incorporated into training, the annotation requirement will be small. A large part of active learning research is focused on designing suitable metrics to quantify the information contained in an unlabeled sample, that can be used for querying samples from the data pool. A common thread in these works is the notion that choosing samples that confuse the machine the most would result in a better (efficient) active learning performance.
	\begin{figure}[h]
		\centering
		\includegraphics[width=0.75\textwidth]{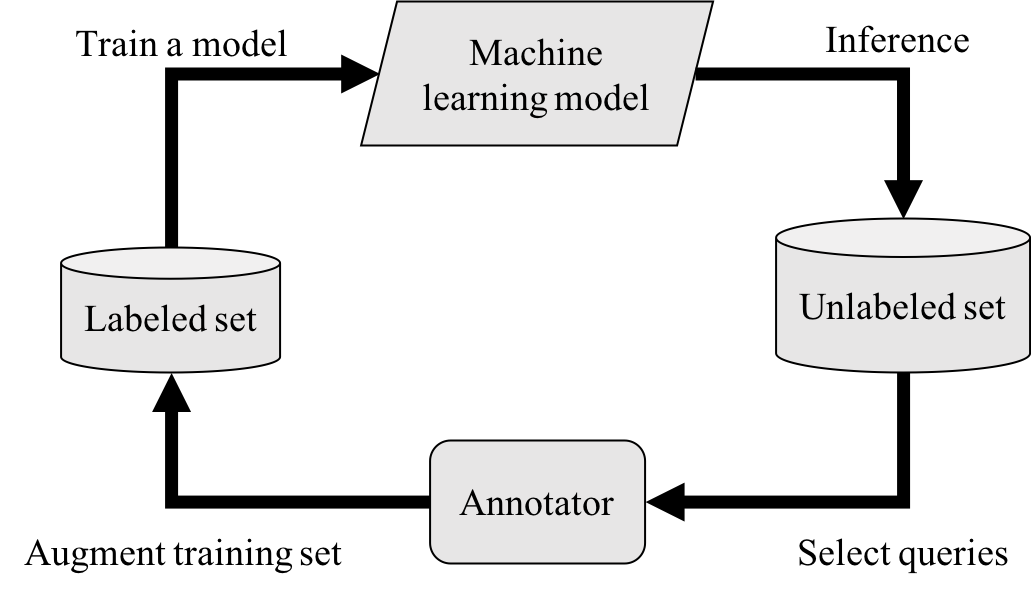}
		\caption{Illustration of an active learning system.}
		\label{fig:advanDL-al}
	\end{figure}
	
	Active learning with deep neural networks has obtained increasing attention within the remote sensing community in recent years~\cite{sun2016active,liu2017active,deng2018active,lin2018active,haut2018active}. Liu \etal\ ~\cite{liu2017active} used features produced by a DBN to estimate the representativeness and uncertainty of samples. Both~\cite{deng2018active} and~\cite{lin2018active} explored using an active learning strategy to facilitate transferring knowledge from one dataset to another. In~\cite{deng2018active}, a stacked sparse autoencoder was initially trained in the source domain and then fine-tuned in the target domain. To overcome the labeled data bottleneck, an uncertainty-based metric was used to select the most informative samples from the source domain for active learning. Similarly, Lin \etal\ ~\cite{lin2018active} trained two separate autoencoders from the source and target domains. Representative samples were selected based on the density in the neighborhood of the samples in the feature space. This allowed autoencoders to be effectively trained using limited data. In order to transfer the supervision from source to target domain, features in both domains were aligned by maximizing their correlation in a latent space.
	
	Unlike autoencoders and DBN, convolution neural networks (CNNs) provide an effective framework to exploit the spatial correlation between pixels in a hyperspectral image. However, when it comes to training with small data, CNNs tends to overfit due to the large number of trainable network parameters. To solve this problem,  Haut~\cite{haut2018active} present an active learning algorithm that uses a special network called Bayesian CNN~\cite{gal2015bayesian}. {Gal and Ghahramani~\cite{gal2015bayesian} have shown that dropout in neural network can be considered as an approximation to the Gaussian process, which offers nice properties such as uncertainty estimation and robustness to overfitting. By performing dropout after each convolution layer, the training of Bayesian CNN can be cast as approximate Bernoulli variational inference. During evaluation, outputs of a Bayesian CNN are averaged over several passes, which allows us to estimate the model prediction uncertainty and the model suffers less from overfitting. Multiple uncertainty-based query criteria were then deployed to select samples for active learning.}
	
	\section{Knowledge transfer between sources}
	\label{advanDL-knowtrans}
	Another common image analysis scenario entails learning with multiple sources, in particular where one source is label ``rich'' (in the quantity and/or quality of labeled data), and the other source is label ``starved''. Sources in this scenario could imply different sensors, different sensing platforms (e.g. ground-based imagers, drones or satellites), different time-points and different imaging view-points. In this situation, when it is desired to undertake analysis in the label starved domain (often referred to as the target domain), a common strategy is to transfer knowledge from the label rich domain (often referred to as the source domain). 
	
	\subsection{Transfer learning and Domain Adaptation}
	\label{advanDL-trans}
	Effective training has always been a challenge with deep learning models. Besides requiring large amounts of data, the training itself is time-consuming and often comes with convergence and generalization problems. One major breakthrough of effective training of deep networks is the pretraining technique introduced by Hinton \etal\ in~\cite{hinton2006fast}, where a DBN was pretrained with unlabeled labeled data in a greedy layer-wise fashion, followed by a supervised fine-tuning. {In particular, the DBN was trained one layer at a time by reconstructing outputs from the previous layer for the unsupervised pretraining. At the last training stage, all parameters were fine-tuned together by optimizing a supervised training criterion.} In Erhan \etal\ in~\cite{erhan2010does}, the authors suggested that unsupervised pretraining works as a form of regularization. It not only provides a good initialization but also helps the generalization performance of the network. Similar to unsupervised pretraining, networks pretrained with supervision have also achieved huge success. Infact, using pretrained models as a starting point for new training has become a common practice for many analysis tasks~\cite{sermanet2013overfeat,donahue2014decaf}. 
	
	The main idea behind transfer learning is that knowledge gained from related tasks or a related data source can be transferred to a new task by fine-tuning on the new data. This is particular useful when there is a data shortage in the new domain. In the computer vision community, a common approach to transfer learning is to initialize the network with weights that are pretrained for image classification on the ImageNet dataset~\cite{deng2009imagenet}. The rationale for this is that ImageNet contains millions of natural images that are manually annotated and models trained with it tend to provide a ``baseline performance' with generic and basic features commonly seen in natural images. Researchers have shown that features from lower layers of deep networks are color blobs, edges, shapes ~\cite{yosinski2015understanding}. These basic features are usually readily transferable across datasets (e.g. data from different sources) ~\cite{yosinski2014transferable}. 
	
	In~\cite{penatti2015deep}, Penatti \etal\ discussed the feature generalization in the remote sensing domain. Empirical results suggested that transferred features are not always better than hand-crafted features, especially when dealing with unique scenes in remote sensing images. Windrim \etal\ ~\cite{windrim2018pretraining} unveiled valuable insights of transfer learning in the context of hyperspectral image classification. In order to test the effect of filter size and wavelength interval, multiple hyperspectral datasets were acquired with different sensors. The performance of transfer learning was examined through a comparison with training the network from scratch, i.e., randomly initializing network weights. Extensive experiments were carried out to investigate the impact of data size, network architecture, and so on. The authors also discussed the training convergence time and feature transferability under various conditions.
	
	Despite the open questions which requires more investigations, extensive studies have empirically shown the effectiveness of transfer learning on hyperspectral image analysis~\cite{marmanis2016deep,zhang2016weakly,yang2017learning,mei2017learning,othman2017domain,yuan2017hyperspectral,shi2017can,tao2017unsupervised,ma2018super,liu2018classifying,niu2018weakly,sumbul2018fine,zhou2018deep}. Marmanis \etal\ ~\cite{marmanis2016deep}  introduced the pretrained model idea~\cite{krizhevsky2012imagenet} for hyperspectral image classification. A pretrained AlexNet~\cite{krizhevsky2012imagenet} was used as a fixed feature extractor and a two-layer CNN was attached for the final classification. Yang \etal\ ~\cite{yang2017learning} proposed a two-branch CNN for extracting spectral-spatial features. To solve the data scarcity problem, weights of lower layers were pretrained from another data set and the entire network was then fine-tuned on the source dataset. Similar strategies have also been followed in~\cite{xie2016transfer,mei2017learning,liu2018classifying}.
	
	Along with pretraining and fine-tuning, domain adaptation is another mechanism to transfer knowledge from one domain to another. Domain adaptation algorithms aim at learning a model from source data that can perform well on the target data. It can be considered as a sub-category of transfer learning, where the input distribution $p(X)$ changes while the label distribution $p(Y|X)$ remains the same across the two domains. Unlike the pretraining and fine-tuning method, which can be used when both distributions change, domain adaptation usually assumes the class-specific properties of the features within the two domains are correlated. This allows us to enforce stronger connections while transferring knowledge.
	
	Othman \etal\ ~\cite{othman2017domain} proposed a domain adaptation network that can handle cross-scene classification when there is no labeled data in the target domain. Specifically, the network used three loss components for training: a classification loss (cross entropy) in the source domain, a domain matching loss based on maximum mean discrepancy (MMD)~\cite{fortet1953mmd}, and a graph regularization loss that aims to retain the geometrical structure of the unlabeled data in the target domain. The cross entropy loss ensures that features produced by the network are discriminative. Having discriminative features in the original domain has also been found to be beneficial for the domain matching process~\cite{zhou2017domain}. In order to undertake domain adaptation, features from the two domains were aligned by minimizing the distribution difference. Zhou and Prasad~\cite{zhou2018deep} proposed to align domains (more specifically, features in these domains) based on domain adaptation transformation learning (DATL)  ~\cite{zhou2017domain} -- DATL aligns class-specific features in the two domains by projecting the two domains onto a common latent subspace such that the ratio of within-class distance to between-class distance is minimized in that latent space. 
	
	Next, we briefly review how a projection such as DATL can be used to align deep networks for domain adaptation and present some results with multi-source hyperspectral data. Consider the distance between a source sample $x^s_i$ and a target sample $x^t_j$ in the latent space,
	\begin{equation}
	\label{eq:distance}
	d(x^s_i, x^t_j) = \| f_s(x^s_i) - f_t(x^t_j) \|^2,
	\end{equation}
	where $f_s$ and $f_t$ are feature extractors, e.g., CNNs, that transform samples from both domains to a common feature space. To make the feature space robust to small perturbations in original source and target domains, the stochastic neighborhood embedding is used to measure classification performance~\cite{hinton2003stochastic}. In particular, the probability $p_{ij}$ of the target sample $x^t_j$ being the neighbor of the source sample $x^s_i$, is given as
	\begin{equation}
	p_{ij} = \frac{\exp(-\| f_s(x^s_i) - f_t(x^t_j)\|^2)}{\sum_{x^s_k \in \mathcal{D}^S} \exp(-\| f_s(x^s_k) - f_t(x^t_j)\|^2) },
	\end{equation}
	where $\mathcal{D}^S$ is the source domain. Given a target sample with its label $(x^t_j, y^t_j = c)$, the source domain $\mathcal{D}^s$ can be split into a \emph{same-class} set $\mathcal{D}^s_c = \{x^s_k| y_k = c\}$ and a \emph{different-class} set $\mathcal{D}^s_{\not c} = \{x^s_k| y_k \neq c\}$. In the classification setting, one wants to maximize the probability of making the correct prediction for $x_j$.
	\begin{equation}
	\label{datl}
	p_{j} = \frac{\sum_{x^s_i \in \mathcal{D}^s_c} \exp(-\| f_s(x^s_i) - f_t(x^t_j)\|^2)}{\sum_{x^s_k \in \mathcal{D}^s_{\not c}} \exp(-\| f_s(x^s_k) - f_t(x^t_j)\|^2)}.
	\end{equation}
	Maximizing the probability $p_j$ is equivalent to minimizing the ratio of intra-class distances to inter-class distances in the latent space. {This ensures that classes from the target domain and the source domain are aligned in the latent space. Note that the labeled data from the target domain (albeit limited) can be further used to make the feature more discriminative. The final objective function of DATL can then be written as}
	\begin{equation}
	\mathcal{L} = 
	\beta \frac{\sum_{x^s_i \in \mathcal{D}^s_c} \exp(-\| f_s(x^s_i) - f_t(x^t_j)\|^2)}{\sum_{x^s_k \in \mathcal{D}^s_{\not c}} \exp(-\| f_s(x^s_k) - f_t(x^t_j)\|^2)} + 
	(1-\beta) \frac{\sum_{x^t_i \in \mathcal{D}^t_c} \exp(-\| f_t(x^s_i) - f_t(x^t_j)\|^2)}{\sum_{x^t_k \in \mathcal{D}^t_{\not c}} \exp(-\| f_t(x^s_k) - f_t(x^t_j)\|^2)}.
	\label{eq:datl-obj}
	\end{equation}
	The first term can be domain alignment term and the second term can be seen as a class separation term. $\beta$ is the trade-off parameter that is data dependent. The greater the difference between source and target data, the larger value of $\beta$ should be used to put more emphasis on domain alignment.
	
	Depending on the feature extractors, Eq.~\ref{datl} can be either solved using conjugate gradient-based optimization~\cite{zhou2017domain} or treated as a loss and solved using stochastic gradient descent~\cite{xu2019d}. DATL has been shown to be effective for addressing large domain shifts such as between street-view and satellite hyperspectral images~\cite{zhou2017domain} acquired with different sensors and imaged with different viewpoints.
	
	Figure~\ref{fig:advanDL-fann} shows the architecture of feature alignment neural network (FANN) that leverages DATL. Two convolutional recurrent neural networks (CRNN) were trained separately for the source and target domain. Features from corresponding layers were connected through an adaptation module, which is composed of a DATL term and a trade-off parameter that balances the domain alignment and the class separation. Specifically, the trade-off parameter $\beta$ is automatically estimated by a proxy A-distance (PAD)~\cite{ben2007analysis}.
	\begin{equation}
	\beta = \text{PAD}/2 =  1 - 2 \epsilon,
	\label{eq:fann-beta}
	\end{equation}
	where PAD is defined as $\text{PAD} = 2(1-2\epsilon)$ and $\epsilon \in [0, 2]$ is the generalization error of a linear SVM trained to discriminate between two domains.
	Aligned features were then concatenated and fed to a final softmax layer for classification.
	\begin{figure}[h]
		\centering
		\includegraphics[width=\textwidth]{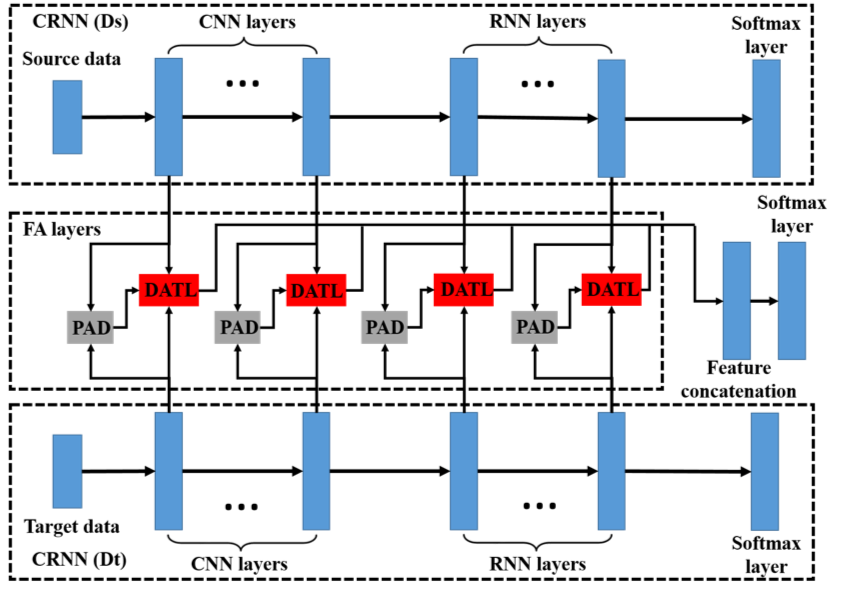}
		\caption{The architecture of feature alignment neural network.  (Source adapted from \cite{zhou2018deep})}
		\label{fig:advanDL-fann}
	\end{figure}
	\begin{table}[h] 
		\centering
		\caption{Network configuration summary for the Aerial and Street view wetland hyperspectral dataset (A-S view wetland). (Source adapted from \cite{zhou2018deep})}
		\begin{tabular}{c}
			\hline
			FANN (A-S view wetland)\\
			\hline
			CRNN (Street) $\rightarrow$ DATL $\leftarrow$  CRNN (Aerial) \\
			\hline
			(conv4-128 + maxpooling) $\rightarrow$ DATL $\leftarrow$ (conv5-512 + maxpooling)  \\
			(conv4-128 + maxpooling) $\rightarrow$ DATL $\leftarrow$ (conv5-512 + maxpooling)  \\
			(conv4-128 + maxpooling) $\rightarrow$ DATL $\leftarrow$ (conv5-512 + maxpooling)  \\
			(conv4-128 + maxpooling) $\rightarrow$ DATL $\leftarrow$ (conv5-512 + maxpooling)  \\
			(conv4-128 + maxpooling) $\rightarrow$ DATL $\leftarrow$ (conv5-512 + maxpooling)  \\
			recur-64 $\rightarrow$ DATL $\leftarrow$ recur-128	\\
			\hline
			fully connected-12\\ 
			\hline
		\end{tabular}
		\label{tab:advanDL-config}
	\end{table}
	\begin{figure}[h]
		\centering
		\includegraphics[width=\textwidth]{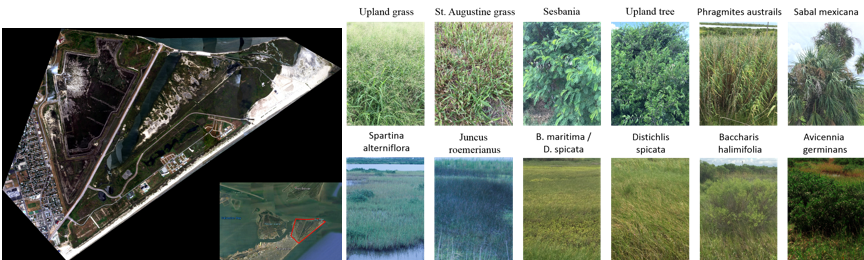}
		\caption{Aerial and Street view wetland hyperspectral dataset. Left: aerial view of the wetland data (target domain). Right: street view of wetland data (source domain). (Source adapted from \cite{zhou2018deep})}
		\label{fig:advanDL-wetland}
	\end{figure}
	\begin{figure}[h]
		\centering
		\includegraphics[width=0.45\textwidth]{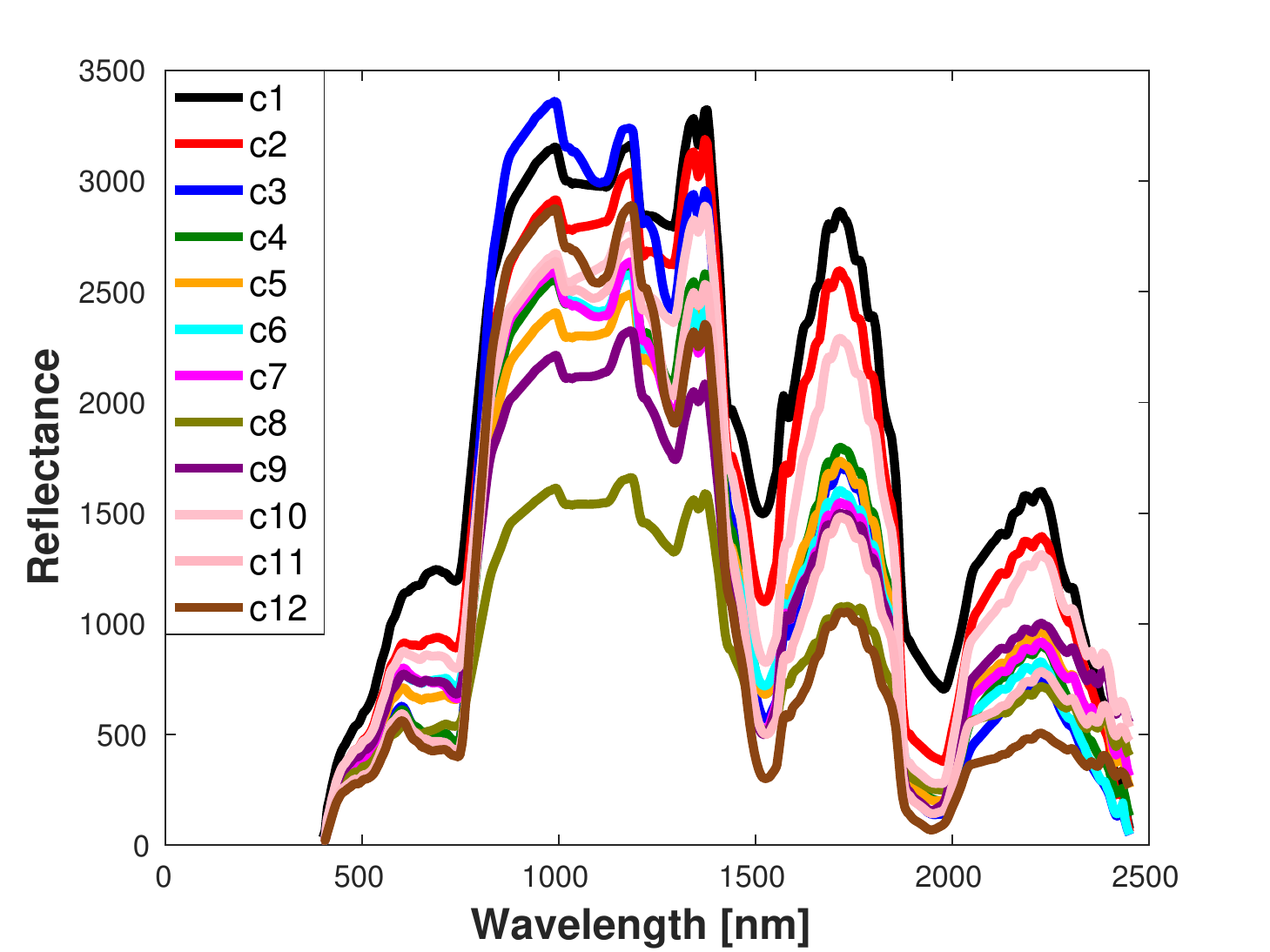}
		\includegraphics[width=0.45\textwidth]{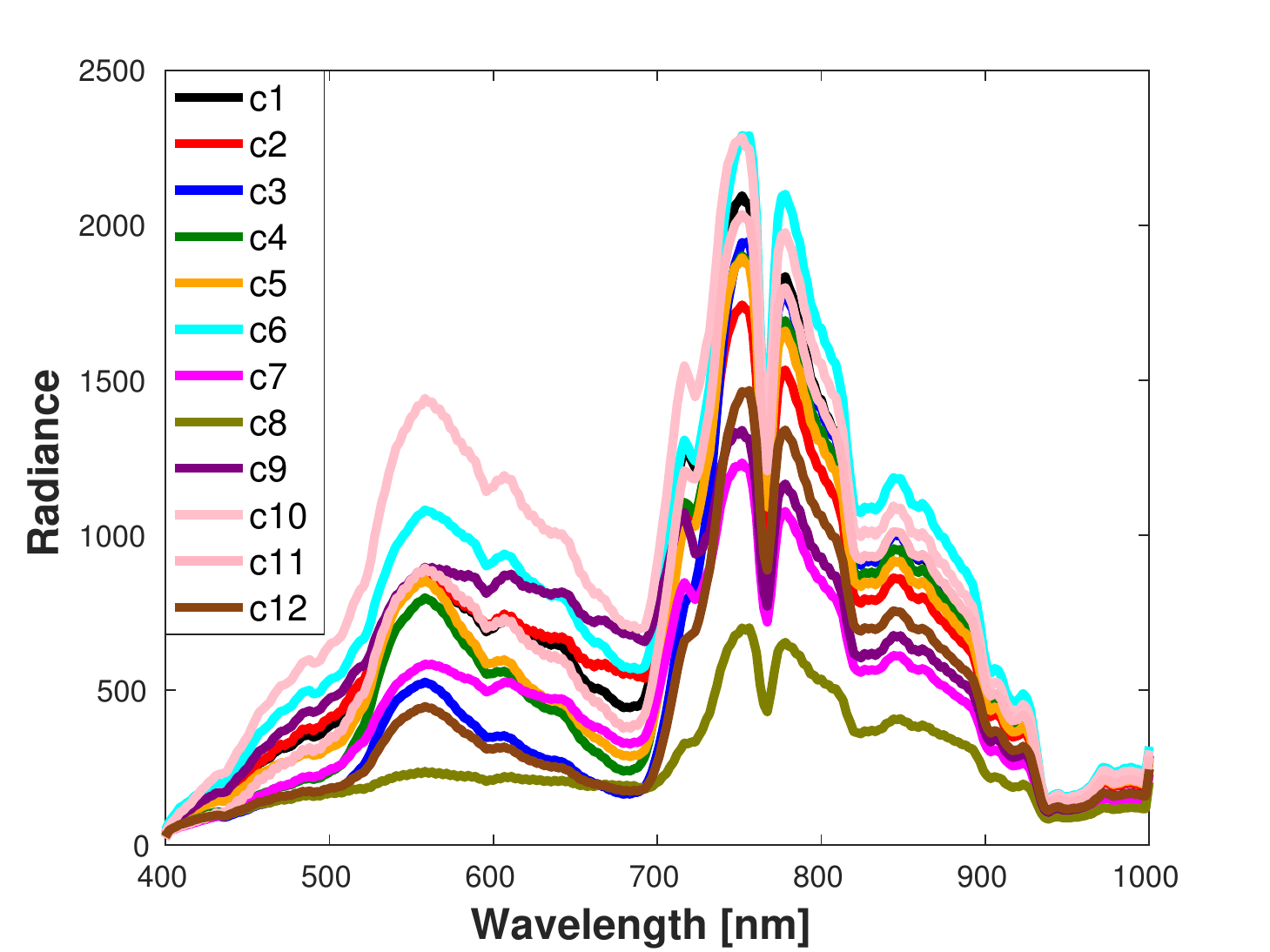}  \\
		\includegraphics[width=0.75\textwidth]{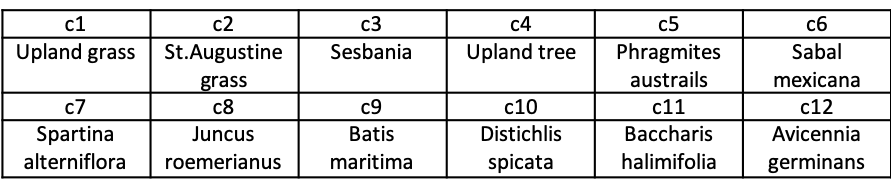}
		\caption{Mean spectral signature of the aerial view (target domain) wetland data (a) and the street view (source domain) wetland data (b). Different wetland vegetation species (classes) are indicated by colors. (Source adapted from \cite{zhou2018deep})}
		\label{fig:advanDL-sigSH}
	\end{figure}
	
	The performance of FANN was evaluated on a challenging domain adaptation dataset introduced in~\cite{zhou2017domain}. See Fig.~\ref{fig:advanDL-wetland} for the true color images for the source and target domains. The dataset consists of hyperspectral images of ecologically sensitive wetland vegetation that were collected by different sensors from two viewpoints -- ``aerial'', and ``street-view'' (and using sensors with different spectral characteristics) in Galveston, TX.  {Specifically, the aerial data were acquired using the ProSpecTIR VS sensor aboard an aircraft, and has 360 spectral bands ranging from 400 nm to 2450 mm with a 5 nm spectral resolution. The aerial view data were radiometrically calibrated and corrected. The resulting reflectance data has a spatial coverage of $3462 \times 5037$ pixels at a 1 m spatial resolution.  On the other hand, the street view data were acquired through the Headwall Nano-Hyperspec sensor on a different date and represents images acquired by operating the sensor on a tripod and imaging the vegetation in the field during ground-reference campaigns. Unlike the aerial view data, the street view data represent at-sensor radiance data with 274 bands spanning 400 nm and 1000 nm at 3 nm spectral resolution.} As can be seen from Fig.~\ref{fig:advanDL-sigSH}, spectral signatures for the same class are very different between the source and target domains. With very limited labeled data in the aerial view, FANN achieved significant classification improvement compared to traditional domain adaptation methods (see Table~\ref{tab:advanDL-fann}). 
	\begin{table}[h]
		\centering
		\begin{tabular}{lcccc}
			\toprule
			Methods & SSTCA & KEMA & D-CORAL & FANN \\
			\midrule
			Accuracy & $85.3 \pm 5.6$ & $87.3 \pm 1.7$ & $92.5 \pm 1.9$ & $95.8 \pm 1.1$ \\
			\bottomrule
		\end{tabular}
		\caption{Overall classification accuracies of different domain adaptation methods on the aerial and street view wetland dataset.  (Source adapted from \cite{zhou2018deep})}
		\label{tab:advanDL-fann}
	\end{table}
	
	\begin{figure}[h]
		\centering
		\includegraphics[width=\textwidth]{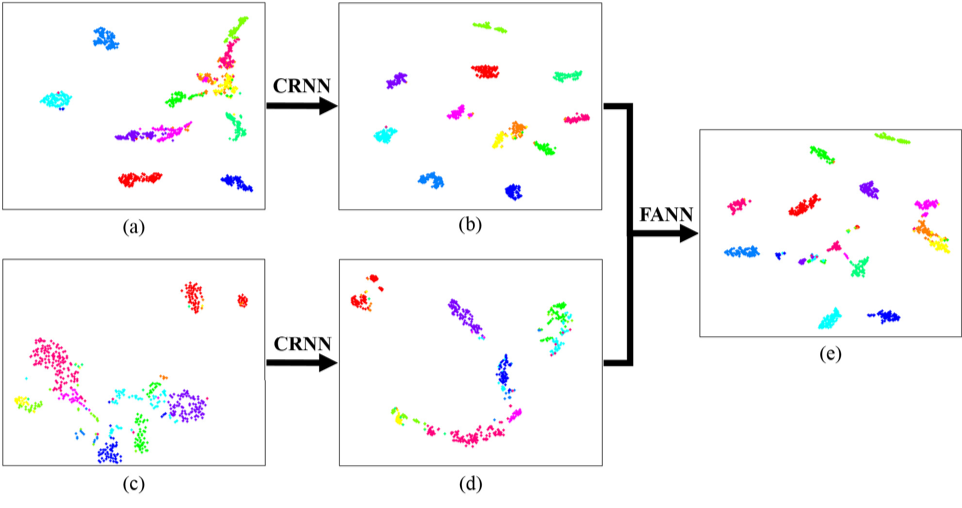}
		\caption{t-SNE feature visualization of the aerial and street view wetland hyperspectral data at different stages of FANN. (a) Raw spectral features of street view data in source domain. (b) CRNN features of street view data in source domain. (c) Raw spectral features
			of aerial view data in source domain. (d) CRNN features of aerial view data in source domain. (e) FANN features for both domains in the latent space.  (Source adapted from \cite{zhou2018deep})}
		\label{fig:advanDL-fann-tsne}
	\end{figure}
	As can be seen from Fig.~\ref{fig:advanDL-fann-tsne}, raw hyperspectral features from source and target domains are not aligned with each other. Due to the limited labeled data in aerial view data, the mixture of classes happens in a certain level. The cluster structures are improved slightly by CRNN, see Fig.~\ref{fig:advanDL-fann-tsne} (c) and (d) for comparison. On the contrary, the source data, i.e., street view data, have well-separated cluster structure. However, the classes are not aligned between the two domains, therefore, labels from the source domain cannot be used to directly train a classifier for the target domain. After passing all samples through the FANN, the two domains are aligned class-by-class in the latent space, as shown in Fig.~\ref{fig:advanDL-fann-tsne} (e).
	
	\begin{table}[h]
		\centering
		\caption{Overall accuracy of the features of alignment layers and concatenated features for the Aerial and Street view wetland dataset. (Source adapted from \cite{zhou2018deep})}
		\begin{tabular}{cccccccc}
			\toprule
			\textit{\textbf{Layer}} &  \textit{\textbf{FA-1}} &  \textit{\textbf{FA-2}} &  \textit{\textbf{FA-3}} &  \textit{\textbf{FA-4}} &  \textit{\textbf{FA-5}}  &  \textit{\textbf{FA-6}} &  \textit{\textbf{FANN}}  \\
			\midrule
			\textit{\textbf{OA}} &88.1 &86.2 &83.9 &75.7 &72.0 &86.4 &95.8	\\
			\bottomrule
		\end{tabular}
		\label{tab:advanDL-layer_acc}
	\end{table}
	{To better understand the feature adaptation process, features from all layers were investigated individually and compared to the concatenated features. Performance of each alignment layer are shown in Table~\ref{tab:advanDL-layer_acc}. Consistent with observations in~\cite{yosinski2014transferable}, accuracies drop from the first layer to the fifth layer as features become more and more specialized toward the training data. Therefore, the larger domain gap makes domain adaptation challenging. Although the last layer (FA-6) was able to mitigate this problem, this is because the recurrent layer has the ability to capture contextual information along the spectral direction of the hyperspectral data. Features from the last layer are the most discriminative ones, which allow the aligning module (DATL) to put more weight on the domain alignment (c.f. $\beta$ in Eq.~\ref{eq:datl-obj} and Eq.~\ref{eq:fann-beta}). The concatenated features obtained the highest accuracy compared to individual layers. As mentioned in~\cite{zhou2018deep}, an improvement of this idea would be to learn a combination weights for different layers instead of a simple concatenation.}
	
	\subsection{Transferring Knowledge -- Beyond Classification}
	\label{advanDL-od}
	In addition to image classification / semantic segmentation tasks, the notion of transferring knowledge between sources and datasets has also been used for many other tasks, such as object detection~\cite{zhang2016weakly}, image super-resolution~\cite{yuan2017hyperspectral}, and image captioning~\cite{shi2017can}.
	
	Compared to image-level labels, training an object detection model requires object-level labels and corresponding annotations (e.g. through bounding boxes). This increases the labeling requirement/costs for efficient model training. Effective feature representation is hence crucial to the success of these methods. As an example, in order to detect aircraft from remote sensing images, Zhang \etal\ ~\cite{zhang2016weakly} proposed to use the UC Merced land use dataset~\cite{yang2010bag} as a background class to pretrain Faster RCNN~\cite{ren2015faster}. By doing this, the model gained an understanding of remote sensing scenes which facilitated robust object detection. The underlying assumption in such an approach is that even though the foreground objects may not be the same, the background information remains largely unchanged across the sources (e.g. datasets), and can hence be transferred to a new domain. 
	
	Another important application of remote sensed images is pansharpening, where a panchromatic image (which has a coarse/broad spectral resolution, but very high spatial resolution) is used to improve the spatial resolution of multi/hyperspectral image. However, a high-resolution panchromatic image is not always available for the same area that is covered by the hyperspectral images. To solve this problem, Yuan \etal\ ~\cite{yuan2017hyperspectral} pretrained a super-resolution network with natural images and applied the model to the target hyperspectral image band by band. The underlying assumption in this work is that the spatial features in both the high- and low-resolution images are the same in both domains irrespective of the spectral content.
	
	Traditional visual tasks like classification, object detection, and segmentation interpret an image at either pixel or object level. Image captioning takes this notion a step further and aims to summarize a scene in a language that can be interpreted easily. Although many image captioning methods have been proposed for natural images, this topic has not been rigorously developed in the remote sensing domain. Shi \etal\ ~\cite{shi2017can} proposed satellite image captioning by using a pretrained fully convolutional network (FCN)~\cite{long2015fully}. The base network was pretrained for image classification on ImageNet. To understand the images, three losses were defined at the object, environment, and landscape scale respectively. Predicted labels at different levels were then sent to a language generation model for captioning. In this work, the task in target domain is very different from the one in the source domain. Despite that, pretrained model still provides features that are generic enough to help understanding the target domain images.
	
	
	\section{Data augmentation}
	\label{advanDL-aug}
	Flipping and rotating images usually does not affect the class labels of objects within the image. A machine learning model can benefit if the training library is augmented with samples with these simple manipulations. By changing the input training images in a way that does not affect the class, it allows algorithms to train from more examples of the object, and the models hence generalize better to test data. Data generation and augmentation share the same philosophy -- to generate synthetic or transformed data that is representative of real-world data and can be used to boost the training. 
	
	Data augmentation such as flipping, rotation, cropping and color jittering have been shown to be very helpful for training deep neural networks~\cite{krizhevsky2012imagenet,liu2016ssd,chen2017deeplab}. These operations infact have become common practice when training models for natural image analysis tasks. 
	Despite the differences between hyperspectral and natural images, standard augmentation methods like rotation, translation and flipping have been proven to be useful in boosting the classification accuracy of hyperspectral image analysis tasks \cite{lee2017going} and \cite{yu2017deep}. {To simulate the variance in the at-sensor radiance and mixed pixels during the imaging process, Chen \etal\ ~\cite{chen2016deep} created \emph{virtual samples} by multiplying random factors to existing samples and linearly combining samples with random weights respectively.} Li \etal\ ~\cite{li2019data} showed the performance can be further improved by integrating spatial similarity through pixel block pairs, {in which a $3 \times 3$ window around the labeled pixel was used as a block and different blocks were paired together based on their labels to augment the training set.} A similar spatial constraint was also used by Feng \etal\ ~\cite{feng2019cnn}, where {unlabeled pixels were assigned labels for training if their $k$-nearest neighbors (in both spatial and spectral domains) belong to the same class}. Haut \etal\ ~\cite{haut2019hyperspectral} used a random occlusion idea to augment data in the spatial domain. It randomly erases regions from the hyperspectral images during the training. As a consequence, the variance in the spatial domain increased and led to a model that generalized better.
	
	Some flavors of data fusion algorithms can be thought of as playing the role of data augmentation, wherein supplemental data sources are helping the training of the models. For instance, a building roof and a paved road both can be made from similar materials -- in such a case, it may be difficult for a model to tell differential these classes from the reflectance spectra alone. However, this distinction can be easily made by comparing their topographic information (e.g. using LiDAR data).  {A straightforward approach to fuse hyperspectral and LiDAR data would be training separate networks -- one for each source/sensor and combining their features either through concatenation~\cite{xu2017multisource,li2018hyperspectral} or some other schemes such as a composite kernel~\cite{feng2019multisource}. Zhao \etal\ ~\cite{zhao2017superpixel} presented data fusion of multispectral and panchromatic images. Instead of applying CNN to the entire image, features were extracted for superpixels that were generated from the multispectral image. Particularly, a fixed size window around each superpixel was split into multiple regions and the image patch in each region was feed into a CNN for extracting local features. These local features were sent to an auto-encoder for fusion and a softmax layer was added at the end for prediction. Due to its relatively high spatial resolution, the panchromatic image can produce spatial segments at a finer scale than the multispectral image. This was leveraged to refine the predictions by further segmenting each superpixel based on panchromatic image.}
	
	Aside from augmenting the input data, generating synthetic data that resembles real-life data is another approach to increase training samples. Generative adversarial network (GAN) ~\cite{goodfellow2014generative} introduced a trainable approach to generate new synthetic samples. GAN consists of two sub-networks, a generator and a discriminator. During the training, two components play a game with each other. The generator is trying to fool the discriminator by producing samples that are as realistic as possible, and the discriminator is trying to discern whether a sample is synthetically generated or belongs to the training data. After the training process converges, the generator will be able to produce samples that look similar to the training data. Since it does not require any labeled data, there has been an increasing interest in using GAN for data augmentation in many applications. This has recently been applied to the hyperspectral image analysis in recent years~\cite{he2017generative,ma2018super,zhan2018semisupervised,zhu2018generative}. Both~\cite{he2017generative} and~\cite{zhu2018generative} used GAN for hyperspectral image classification, where a softmax layer was attached to the discriminator. Fake data were treated as an additional class in the training set. Since a large amount of unlabeled was used for training the GAN, the discriminator became good at classifying all samples. A transfer learning idea was proposed for super-resolution in ~\cite{ma2018super}, where a GAN is pretrained on a relatively large dataset and fine-tuned on the UC Merced land use dataset~\cite{yang2010bag}.

	\section{Future Directions}
	In this chapter, we reviewed recent advances in deep learning for hyperspectral image analysis. Although a lot of progress has been made in recent years, there is still a lot of open problems, and related research opportunities. In addition to making advances in algorithms and network architectures (e.g. networks for multi-scale, multi-sensor data analysis, data fusion, image super-resolution etc.), there is a need for addressing fundamental issues that arise from insufficient data and the fundamental nature of the data being acquired. Towards this end, the following directions are suggested
	\begin{itemize}
		\item Hyperspectral ImageNet. We have witnessed the immense success brought about in part by the ImageNet dataset for traditional image analysis. The benefit of building a similar dataset for hyperspectral image is compelling. If such libraries can be created for various image analysis tasks (e.g. urban land-cover classification, ecosystem monitoring, material characterization etc.), they will enable learning truly deep networks that learn highly discriminative spatial-spectral features.  
		\item Interdisciplinary collaboration. Developing an effective model for analyzing hyperspectral data requires a deep understanding of both the properties of the data itself and machine learning techniques. With this in mind, networks that reflect the optical characteristics of the sensing modalities (e.g. inter-channel correlations) and variability caused in acquisition (e.g. varying atmospheric conditions) should add more information for the underlying analysis tasks compared to ``black-box'' networks.
	\end{itemize}
	
\bibliographystyle{elsarticle-num}
\bibliography{DL-ref}


\end{document}